\documentclass[11pt]{article}
\usepackage{emnlp2021}  
\usepackage{times}
\usepackage{latexsym}
\usepackage[T1]{fontenc}
\usepackage[utf8]{inputenc}
\usepackage{microtype}

\usepackage{url}

\usepackage{amsfonts}
\usepackage{booktabs}
\usepackage{multirow}
\usepackage{amsmath}
\usepackage{hhline}
\usepackage{colortbl}
\usepackage{enumerate}
\usepackage[normalem]{ulem}
\usepackage{lipsum}  
\usepackage{fdsymbol}
\usepackage{tikz,pgfplots,pgfplotstable}
\usepackage[linecolor=orange]{todonotes}
\usepackage{caption}
\usepackage{tikz}
\usepackage{siunitx}   

\usepackage{ctable}

\definecolor{MengGreen}{rgb}{0.0, 0.5, 0.0}

\usepackage{setspace}

\usepackage{xspace}
\newcommand{\fiscu}{\texttt{LiSCU}\xspace}

\usepackage{enumitem, kantlipsum}

\usepackage{enumitem} 
\newlist{todolist}{itemize}{2}
\setlist[todolist]{label=$\square$}

\usepackage{soul}
\usepackage{color}
\definecolor{light-gray}{gray}{0.9}

\renewcommand{\arraystretch}{0.75}

\title{\textit{``Let Your Characters Tell Their Story''}: \\ A Dataset for Character-Centric Narrative Understanding}

\author{Faeze Brahman$^{1}$ \qquad Meng Huang$^{2}$ \qquad Oyvind Tafjord$^{3}$ \\ \bf Chao Zhao$^{5}$ \qquad Mrinmaya Sachan$^{4}$ \qquad \bf Snigdha Chaturvedi$^{5}$ \\ $^{1}$University of California, Santa Cruz, $^2$University of Chicago \\ $^{3}$Allen Institute for AI, $^4$ETH Zurich, $^{5}$UNC Chapel Hill \\
\smallskip\\\vspace{-0.65cm}\\
\texttt{fbrahman@ucsc.edu,  huangme@uchicago.edu} }

\date{}

\begin{document}
\maketitle

\begin{abstract}
When reading a literary piece, readers often make inferences about various characters' roles, personalities, relationships, intents, actions, etc.
While humans can readily draw upon their past experiences 
to build such a character-centric view of the narrative, \textit{understanding} characters in narratives can be a challenging task for machines. To encourage research in this field of character-centric narrative understanding, we present \fiscu \ -- a new dataset of literary pieces and their summaries paired with descriptions of characters that appear in them. We also introduce two new tasks on \fiscu
: \textit{Character Identification} and \textit{Character Description Generation}. Our experiments with 
several pre-trained language models adapted for these tasks demonstrate that there is a need for better models of narrative comprehension.\footnote{Data and code are available at: \url{https://github.com/fabrahman/char-centric-story}}



\end{abstract}

\section{Introduction}
\label{sec:intro}


\begin{figure}[t]
\centering
\includegraphics[width=1.0\columnwidth]{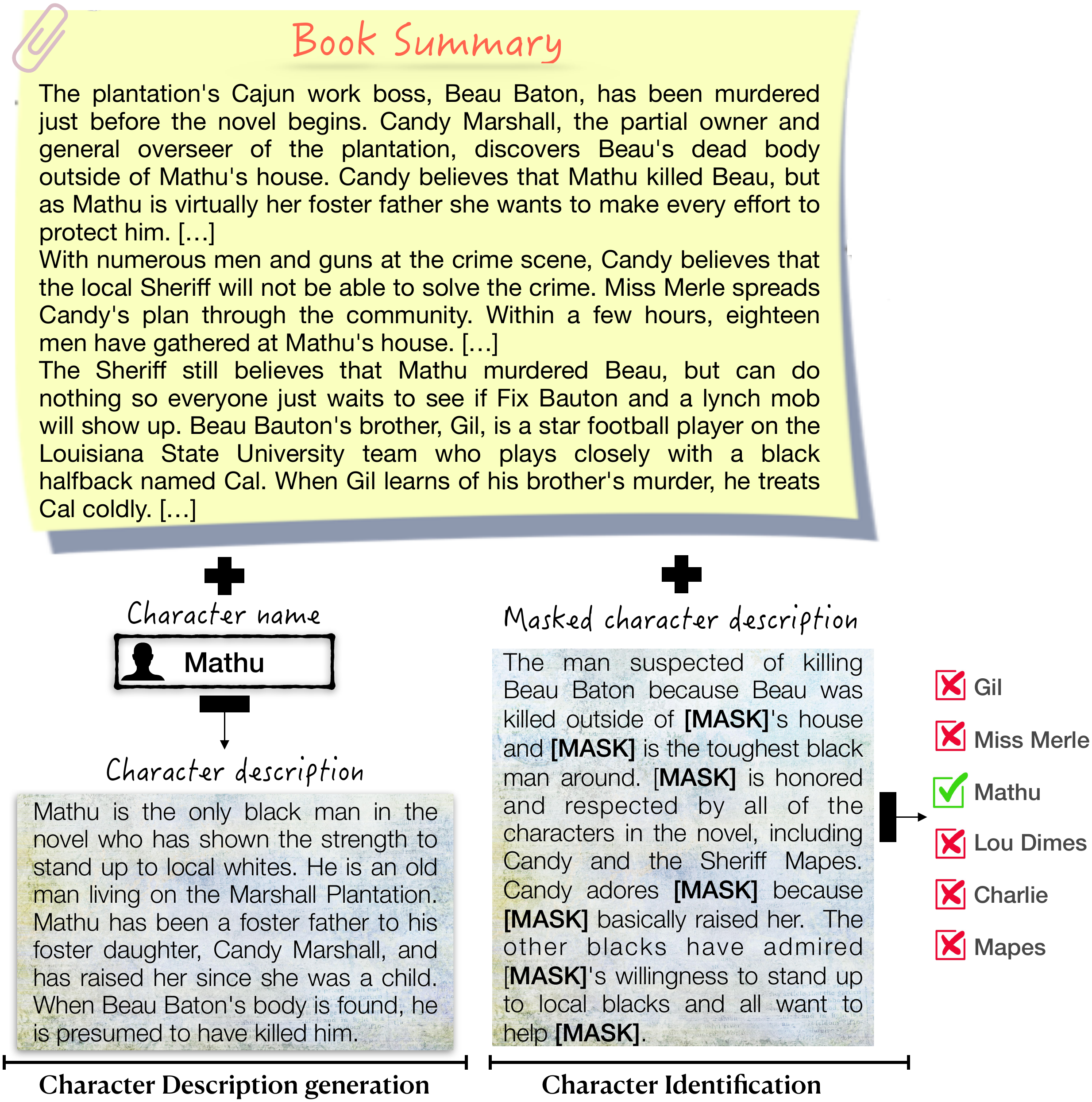} 
\caption{An illustration of the proposed dataset and the two tasks: \textit{Character Description Generation} and \textit{Character Identification}.}
\label{fig:task}
\end{figure}


Previous works in literary analysis have discussed that the development of the plot and the main character(s) are among the most important components that contribute to a good piece of fiction~\cite{kennedy:83, card:99}. In particular, \textit{character(s)} are central to  narratives since their motivations, traits, and actions determine the flow of the plot. 
Hence, understanding and critically analyzing characters is an important facet of literary scholarship.

In 
Computational Narratives, prior work has exploited the potential of character-centric 
natural language understanding~\cite{chambers-2013-event, ChaturvediID17, chu-etal-2018-learning,ZhangCO19}. However, these works are limited to only understanding certain aspects of characters and do not do an in-depth and systematic study.





% ...

To facilitate character-centric narrative understanding, 
we present \fiscu \ -- a new dataset in English, of literary pieces and their summaries paired with descriptions of characters that appear in them. These descriptions analyze the narrative from the perspective of the character highlighting their salient attributes, their role and contribution to the development of the narrative's plot. 


Using this dataset, we devise two new tasks: (1) a \textit{Character Identification} task to identify the character's name from an anonymized character description given the literature summary; and (2) a \textit{Character Description Generation} task to generate the description for a given character of a literature summary. 
Our primary task, \textit{Character Description Generation}, is related, but not identical to summarization. There are two main differences. Summarization typically has a one-to-one correspondence 
between documents and summaries, and
focuses on copying (either extractively or abstractively) important content from the documents to create the summaries. On the other hand, character descriptions are analysis, not merely summaries, of narratives from the character's point of view. They are created by abstracting out the low-level content of the narrative instead of simply identifying and paraphrasing important details. They describe events, roles, relationships, and salient attributes of the character that can be inferred from the narrative and might not be directly stated in the text. In particular, if the narrative describes several events where a character helps the protagonist, the character description will not simply mention all those events, but will instead describe the character as a helpful person (attribute) and a good friend of the protagonist (role). For example, in Fig.~\ref{fig:task}, ``Mathu is virtually her foster father.'' in summary is expressed as ``Candy adores Mathu because he basically raised her.'' in the character description. Thus, the \textit{Character Description Generation} task, provides a unique opportunity for NLP systems to learn to \textit{abstract} and model long-range dependency instead of simply \textit{extracting} information. 

Apart from this novel \textit{abstraction} task, the dataset also poses another challenge for NLP systems by requiring them to process long documents. The average number of tokens in our summaries are $1022$
which is beyond the comfort level of most existing systems. Understanding long narratives and modeling long contexts are new frontiers for NLP research~\cite{RoySVG21,fan2021addressing} and \fiscu pushes us in this direction. To further facilitate research in this direction, we also release a small dataset where the goal is to read the entire literary piece and generate character descriptions.



We explore the ability of the modern neural models on both 
tasks. We demonstrate through experiments that although existing models can identify characters reasonably well in masked descriptions, there is still a scope for improvement considering human accuracy on this task. Also, while existing models can generate fluent and logically self-consistent text, they are not always faithful to the literature summaries and fail to capture salient details about the characters. Our contributions are:
\begin{itemize}[noitemsep,topsep=0.3pt]
    \setlength\itemsep{0.3em}
    \item A new dataset of literature summaries paired with character descriptions to enable character-centric narrative understanding.
    \item A comprehensive human study to assess the quality of the proposed dataset.
    \item Novel tasks: a classification and an abstractive generation task to better understand characters in the narrative plot.
    \item Experiments with several strong baselines and a thorough qualitative analysis.
\end{itemize}


\section{Background}
\label{sec:background}

The field of computational narrative understanding studies how to algorithmically represent,
understand, and generate stories.
Early computational studies on narratives had focused on learning procedural scripts and event sequences~\cite{schank:77a, flairs/ManshadiSG08, regneri-etal-2010-learning}, narrative chains or schemas~\cite{chambers-jurafsky-2008-unsupervised, chambers-jurafsky-2009-unsupervised}, and plot units~\cite{goyal-etal-2010-automatically, mcintyre-lapata-2010-plot, elsner-2012-character}.

Computational linguists have also worked on character-centric modelling of narratives
~\cite{chambers-2013-event}. 
The character-centric perspective aims to understand characters -- their personas, roles, goals, relationships, emotions, etc. 
Previous works have proposed 
methods to detect characters and infer latent personas in movie plot summaries and fictional novels~\cite{bamman-etal-2013-learning, bamman-etal-2014-bayesian,vala:2015,flekova2015personality}, model inter-character relationships~\cite{iyyer:2016, SrivastavaCM16, ChaturvediID17, KimK19}, and emotions~\cite{brahman-chaturvedi-2020-modeling}. 
Earlier works have also considered constructing social networks of characters~\cite{agarwal-etal-2014-frame} from novels~\cite{ElsonDM10, elsner-2012-character} and films~\cite{krishnan-eisenstein-2015-youre}. 




Another line of work related to ours is on summarization of novels~\cite{mihalcea-ceylan-2007-explorations}. This work built a dataset of novel-summary pairs and used unsupervised summarization models 
such as \textit{TextRank}~\cite{mihalcea-tarau-2004-textrank} and \textit{MEAD}~\cite{Radev01experimentsin}. 
Instead of summarizing full novels, \citet{ladhak-etal-2020-exploring} proposed a content-selection approach to create a gold-standard set of extractive summaries by aligning chapter sentences with abstractive summary sentences.

In a more related work, \citet{ZhangCO19} collected a dataset of fictional stories along with author-written summaries. They proposed an extractive ranking and a classification approach to select a subset of salient attributes from a list of candidate attributes 
(extracted from the story) that describe a character's personality. While this work 
presented a collection of personality-related phrases as a potential summary for the actual novel, our dataset contains literature summaries and character descriptions, and we aim to generate natural language texts that analyze the narrative from the perspective of the characters. Such an analysis is more in-depth than a collection of phrases.




\section{The LiSCU Dataset}
\label{sec:data}


We now describe our \textbf{Li}terature \textbf{S}ummary and \textbf{C}haracter \textbf{U}nderstanding (\fiscu) dataset. \fiscu is a dataset of literature summaries paired with descriptions of characters that appear in the summaries. 
Fig.~\ref{fig:task} shows an example of our dataset. 

Next, we describe the data collection pipeline for \fiscu (\S\ref{data-collect}), followed by details on the reproducibility of the data collection process (\S \ref{data_reproducibility}). 

\subsection{Data Collection and Filtering}
\label{data-collect}

We collected \fiscu from various online study guides such as \texttt{shmoop},\footnote{\url{https://www.shmoop.com/study-guides/literature}} \texttt{SparkNotes},\footnote{\url{https://www.sparknotes.com/lit/}}  \texttt{CliffsNotes},\footnote{\url{https://www.cliffsnotes.com/literature}} and \texttt{LitCharts}.\footnote{\url{https://www.litcharts.com}}
These sources contain educational material to help students study for their literature classes. These study guides include summaries of various literary pieces as well as descriptions of characters that appear in them. 
These literature summaries and character descriptions were written by literary experts, typically teachers, and are of high pedagogical quality.

We used \texttt{Scrapy},\footnote{\url{https://scrapy.org/}} a free and open-source web-crawling framework 
to crawl these study guides.
Our initial crawl resulted in a set of $1,774$ literature summaries and $25,525$ character descriptions. These included all characters mentioned in the literary pieces.  
However, not all characters, especially those that played a minor role in the literary piece, appeared in the corresponding literature summaries. Since our task involves making inferences about characters from the literature summaries
, we filtered out the characters which do not appear in the summaries
or their names or the descriptions had very little overlap with the literature summaries.
This is done to mitigate the reference divergence issue~\cite{kryscinski-etal-2019-neural, maynez-etal-2020-faithfulness} and ensure that the literature summary has enough information about the character to generate the description.
For this, we define the ``information overlap'' between two pieces of text $\mathcal{A}$ and  $\mathcal{B}$, $IO(\mathcal{B}||\mathcal{A})$, as the ratio of the length of the longest overlapping word sub-sequence between  $\mathcal{A}$ and  $\mathcal{B}$, over the length of $\mathcal{A}$.\footnote{Technically this is the same as Rouge-L precision} Note that this information overlap measure is not symmetric and intuitively measures how much information about $\mathcal{A}$ is present in $\mathcal{B}$.
We used the information overlap measure to filter our dataset as follows. If the information overlap of the literature summary with the character name, 
$IO($literature summary $||$ character name$)$, is less than $0.6$, then we consider that the character is not prominently mentioned in the literature summary and we remove that character from our dataset. Similarly, if the information overlap between the character description and the literature summary, $IO($literature summary $||$ character description$)$, is less than $0.2$, then we consider the character description generation less feasible and we remove that data point from our dataset.\footnote{These thresholds were chosen by experimenting with different values and manually analyzing the quality of (a subset of) the data.}

\begin{table}[t]
\footnotesize
    \centering

\setlength\tabcolsep{7pt}
\renewcommand{\arraystretch}{1.3}
    \begin{tabular}{l|c}
    \toprule
    \verb|#| unique books & 1,220 \\
    \verb|#| literature summaries & 1,708 \\
    \verb|#| characters & 9,499 \\
    \verb|#| characters with accompanying full book & 2,052 \\
    \verb|#| unique books with full-text & 204 \\
    \midrule
    avg. \verb|#| characters per summary & 5.56 \\
    min. \verb|#| characters per summary & 1 \\
    max. \verb|#| characters per summary & 38 \\
    avg. summary length (in tokens) & 1,022.32 \\
    avg. \verb|#| sentences in summary & 48.82 \\
    avg. character description length (in tokens) & 184.57 \\
    avg. \verb|#| sentences in description & 8.56 \\
    \midrule
    \verb|#| characters in Train set & 7,600 \\
    \verb|#| characters in Test set & 957 \\
    \verb|#| characters in Validation set & 942 \\
    \bottomrule
    \end{tabular}
\setlength\tabcolsep{6pt}

    \caption{Statistics of the \fiscu dataset.}
\label{tab:stat}
\end{table}

However, during these filtering steps, we did not want to remove the most important characters of the narrative. The online study guides list characters in decreasing order of their importance in the literary piece. For example, narrators, protagonists, antagonists, etc., are always described first. Leveraging this ordering, we always retained the top $3$ characters of the literary piece in our dataset.


After the filtering process, our final dataset consists of $1,708$ literature summaries and $9,499$ character descriptions in total. This set was split into train ($80\%$), test ($10\%$), and validation ($10\%$) sets. The data splits were created to avoid any data-leakages -- each literary piece and all of its character descriptions were consistently part of only one of the train, test and validation sets.
Table~\ref{tab:stat} shows the statistics of the final dataset. The dataset also contains the full-text of the books for 
$2,052$ of the character descriptions.





\subsection{Dataset Reproducibility}
\label{data_reproducibility}

\fiscu is drawn from various study guides on the web. While we do not have the rights to directly redistribute this dataset, to allow other researchers to replicate the \fiscu dataset and compare to our work, we provide a simple script that will allow others to recreate \fiscu from a particular time-stamped version of these study guides on \textit{Wayback Machine}, a time-stamped digital archive of the web. Our script  ensures that others will be able to recreate the same train, test and validation splits.

\section{\fiscu Task Definitions}
\label{sec:task-def}
We introduce two new tasks on the \fiscu dataset:
\begin{itemize}[noitemsep,nolistsep]
    \item \textit{Character Identification}
    \item \textit{Character Description Generation}
\end{itemize}


\subsection{Character Identification} 
The \textit{Character Identification} task requires models to identify the character in an anonymized character description.
Given a summary $S$, a candidate list of characters that appear in the literature summary $C=\{c_1, c_2, ..., c_k\}$, and an anonymized character description $D_{masked}^{c*}$,
the goal in this task is to identify the name of the character $c^*$ described in the anonymized character description. We anonymize character descriptions by masking out all mentions of the character $c^*$ in the original description $D^{c*}$. 


\subsection{Character Description Generation} 
The \textit{character description generation} task tests the ability of NLP models to critically analyze the narrative from the perspective of characters and generate coherent and insightful character descriptions.
Formally, given a 
literature summary, $S$, and a character name, $c$, the 
goal in this task is to generate the character's description, $D^{c}$.
Generating the character description necessitates understanding and analyzing every salient information about the character in the literature summary. 

\subsection{Human Assessment of \fiscu}
\label{data-validation}

In order to verify the tractability of these two tasks as well as assessing the quality of the collected \fiscu dataset,
we 
conducted a set of human evaluations on Amazon Mechanical Turk.
We run our human assessment on the full test set of \fiscu. 



\vspace{0.2cm}
\noindent{\bf Assessing the Character Identification task:} In the first human assessment, we showed annotators the literature summaries, anonymized character descriptions 
, and a list of character names 
(plus one randomly sampled character from the literary piece). The descriptions were anonymized by replacing all mentions of the corresponding character names with blanks.\footnote{\label{anonymizing_footnote}We identified mentions of a character in the summary by using 
a coreference system \cite{joshi2019coref,spanbert} as well as by matching the first name or the full name of the character.} For each anonymized character description, we asked 3 judges to identify which character 
it is describing by choosing from the list of 
choices. The judges also had the option of saying that they are unable to identify the character given the literature summary and the anonymized character description.


\begin{figure}[t]
\centering
\includegraphics[width=1\columnwidth]{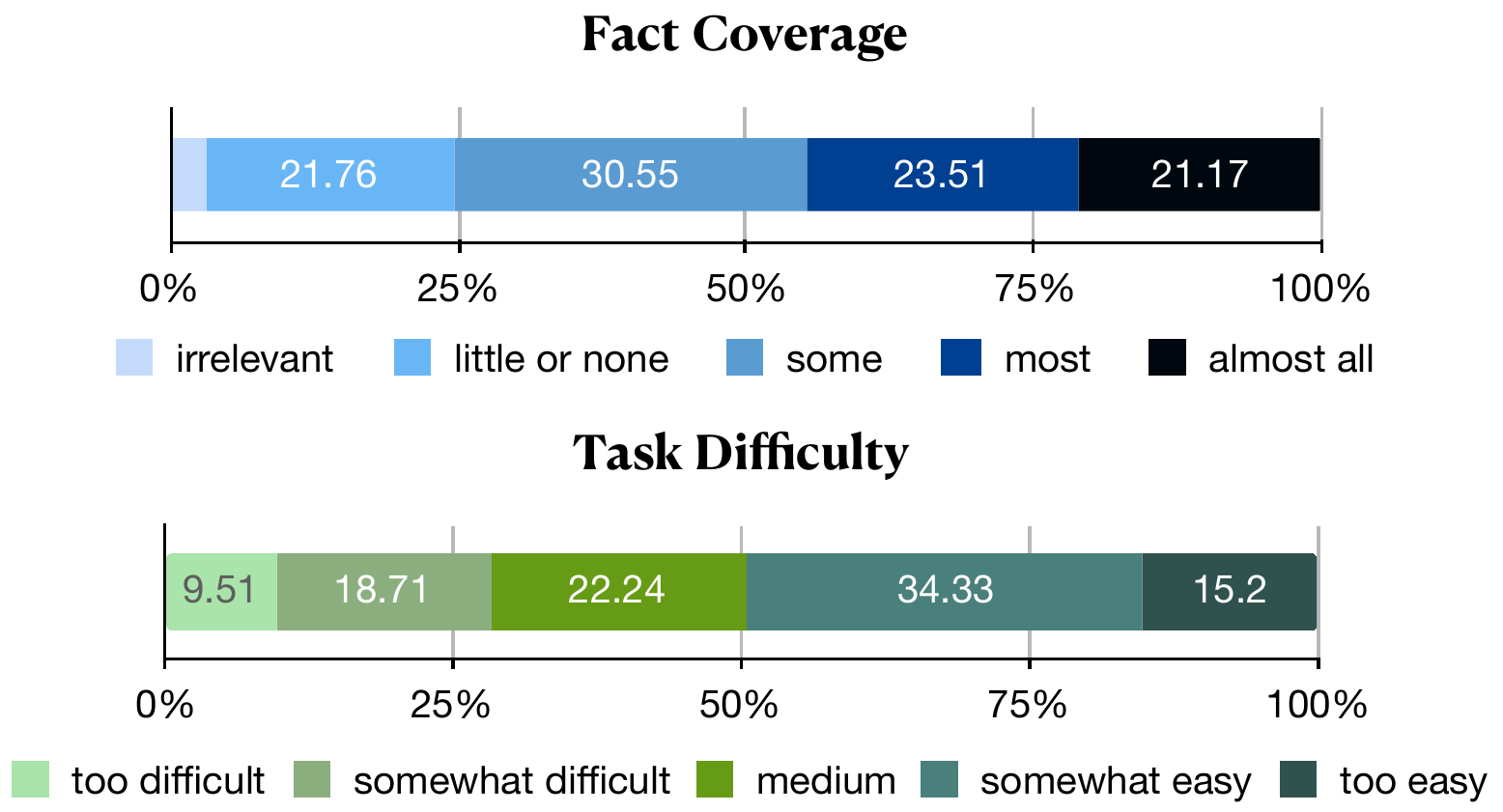} 
\caption{Human assessment of the feasibility of the character description generation task. 
}
\label{fig:human-data-assess}
\end{figure}

\vspace{0.2cm}
\noindent{\bf Assessing the Character Description Generation task:} In the second human assessment, the judges are shown the same summary along with the original de-anonymized character descriptions. For each character description, 3 judges were asked to evaluate the quality of the description by answering the following two questions:

\begin{enumerate}[wide, noitemsep, nolistsep, labelwidth=!, labelindent=0pt]
    \item {\bf Fact coverage:} Specify how much of the information about the specific character in the corresponding ``character description'' is present in the summary (either explicitly or implicitly). Answer choices included: a) \textit{almost all of the information}, b) \textit{most of the information}, c) \textit{some of the information}, d) \textit{little or none of the information}, and e) \textit{character does not appear in the summary at all}.
    \item {\bf Task difficulty:} Given the summary, how easy is it to write the character description on a Likert scale of 0-4 (0 being too difficult, 4 being too easy)? If in the previous question the judges found that some of the information in the character description was not present in the summary, they are asked to disregard that while answering this question. In other words, they only need to consider the information in the character description which is explicitly or implicitly mentioned in the summary. 
\end{enumerate}

\begin{figure}[t!]
\centering
\includegraphics[width=0.95\columnwidth]{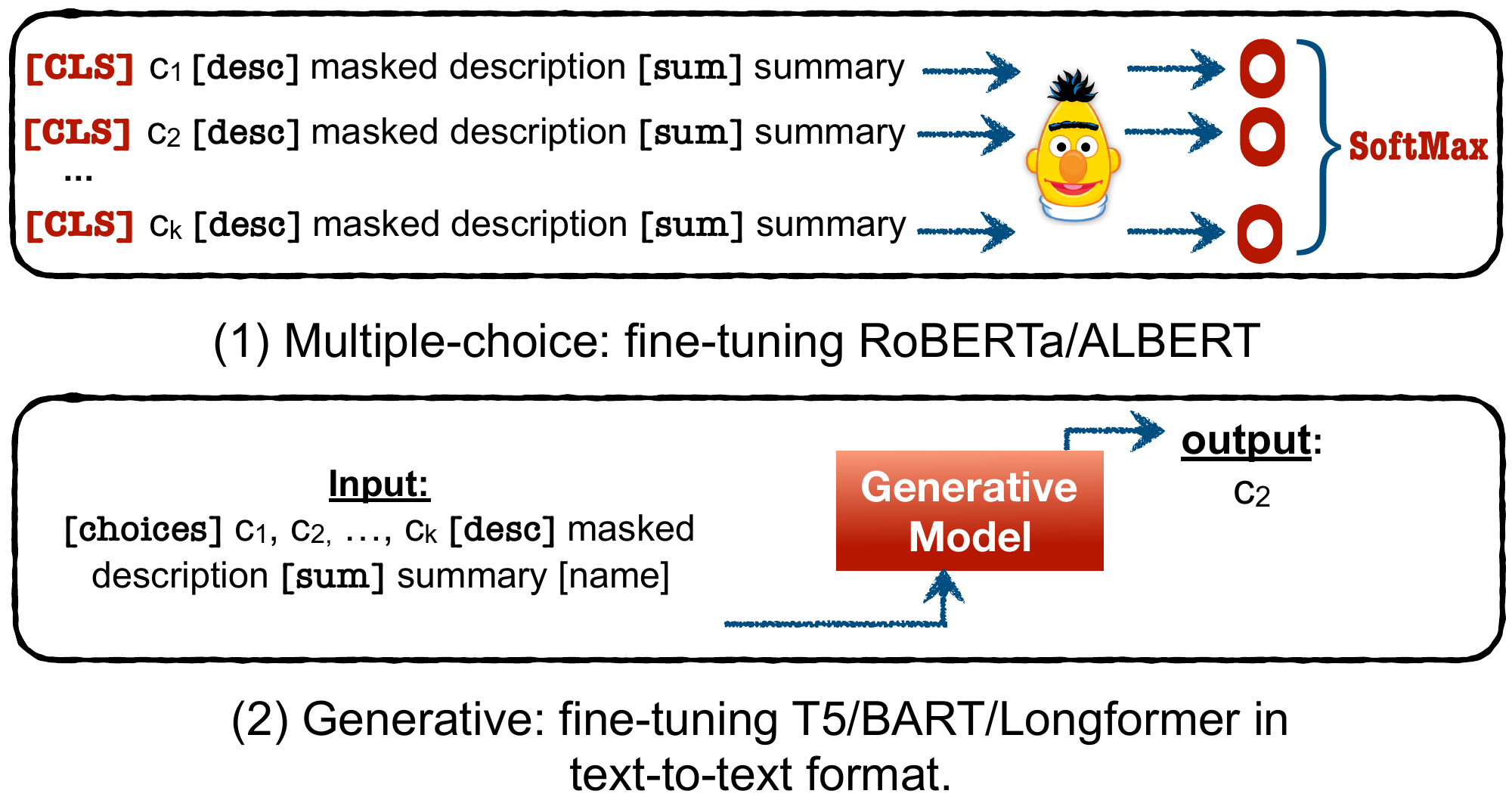}
\caption{Approaches for \textit{Character Identification}.}
\label{fig:identification}
\end{figure}

We recruited $200$ crowd-workers who were located in the US, UK, or CA, and had a $98\%$ approval rate for at least $5,000$ previous annotations. We collected each annotation from $3$ workers and use majority vote in our assessments. In the Appendix~\ref{appendix-annotations}, we describe several steps we took to alleviate limitations of using crowd-sourcing and ensure high quality annotations. Screenshots of our AMT experiments are provided in the Appendix. 

For the first assessment on identifying characters, the human accuracy was $91.80\%$ (Fleiss' Kappa  \cite{landis1977measurement} $\kappa = 0.79 $), indicating the feasibility of the task.

For the second assessment of fact coverage and task difficulty, we summarize the result in Fig.~\ref{fig:human-data-assess}.  
The top chart (`Fact Coverage') shows that 
around 75\% of the of the literature summaries contain reasonable amount of information about the character represented in the corresponding character description. The bottom chart (`Task Difficulty') shows that 
more than 90\% of the times, the human judges considered the task of writing the character descriptions from the literature summaries not too difficult.\footnote{There is a natural label bias in the annotations: most of the responses fell into few categories. In this case, standard inter-annotator agreement statistics are not reliable (the well-known paradoxes of kappa  \cite{feinstein1990high}). Thus, we simply report a pairwise agreement (i.e., how often do two judges agree on the answer for the same question)
of 0.71 and 0.64 for `fact coverage' and `task difficulty', respectively.} 

These results verify the feasibility of understanding and drawing reasonable inferences about characters in the literature summaries from the \fiscu dataset. 
Next, we describe models and establish baseline performances on the two proposed tasks.




\section{Character Identification}


We present two approaches to address this task: (1) solving it as a multiple-choice classification problem, and (2) using a generative classifier that generates, instead of identifying, the character name, as shown in Fig.~\ref{fig:identification}.


In the multiple-choice approach, we use the standard setup introduced in BERT~\cite{devlin-etal-2019-bert} where the text from $c_i$, $D_{masked}^{c*}$ and $S$ (with custom prefix tokens) are concatenated as input, and the 
\texttt{[CLS]} token is projected to a final logit. We apply a Softmax function to the logits to obtain the scores for each $c_i$. For training practicalities, we limit the number of choices to 4 during training (using the earliest window of choices which include the correct one).
During inference, we can generate the logits for all the answer choices since they are independent before the final Softmax.


To establish a baseline performance, we experiment with finetuning RoBERTa~\cite{liu2019roberta}, and ALBERT~\cite{Lan2020ALBERT:} which have been shown to perform well in several classification tasks. 
However, both these models cannot process inputs longer than 512 tokens and the concatenated inputs are generally much longer. So we also tried Longformer~\cite{Beltagy2020LongformerTL}, a BERT-like model with an attention mechanism designed to  scale linearly with sequence length, thus allowing the model to encode longer documents. However, despite trying various hyperparameters, Longformer was not able to match the scores in our experiments. 

\begin{table}[t]
\scriptsize
    \centering
\setlength\tabcolsep{4pt}
\renewcommand{\arraystretch}{1.4}
    \begin{tabular}{lcc}
    \toprule
    \textbf{Model} & \textbf{Description Setup} & \textbf{Accuracy (\%)} \\ 
    \midrule
    \rowcolor{blue!10}\textit{Random Guess} & - & 18.70 \\
    \midrule
    RoBERTa-Large~\cite{liu2019roberta} & partial & 77.84\\
    ALBERT-XXL~\cite{Lan2020ALBERT:} & partial & \textbf{83.33}\\
    \midrule
    T5-11B~\cite{2019t5} & partial & \underline{80.16}\\
    BART-Large~\cite{lewis2019bart} & partial & 74.89\\ 
    Longformer~\cite{Beltagy2020LongformerTL} &  partial & 71.10 \\ 
    BART-Large~\cite{lewis2019bart} & full & 78.58 \\
    Longformer~\cite{Beltagy2020LongformerTL} & full & 74.78\\
    \midrule
    \rowcolor{blue!10} \textit{Human Performance} & - & 91.80 \\ 
    \bottomrule
    \end{tabular}
\setlength\tabcolsep{2pt}
    \caption{Accuracy for the \textit{Character Identification}. The  `partial' description setup used a truncated description ($50$ words) to allow including more of the summary. 
    }
    \label{tab:disc-res}
\end{table}


Our second approach, a generative classifier, is inspired by~\citet{JMLR:v21:20-074} who studied transfer learning by converting NLP problems into a text-to-text format. The generative classifier addresses the character identification problem by directly generating the character name $\hat{c}$, given all character names (answer choices), the masked character description, and the summary (see Fig.~\ref{fig:identification}). 
During inference, we compute the model's probability of each of the answer choices, and output the one with the highest probability.

We use this procedure to train several strong baselines built on top of the following pre-trained transformer-based models: BART~\cite{lewis2019bart}, T5~\cite{JMLR:v21:20-074}, and Longformer~\cite{Beltagy2020LongformerTL}.


\vspace{0.15cm}
\noindent\textbf{Implementation Details.}\space\space The RoBERTa and ALBERT multiple-choice classifiers were trained for 6 epochs, initial learning rate 1e-5 (ADAM optimizer), batch size 16. The generative classifier using BART was trained for 5 epochs, initial learning rate 5e-6, batch size 8. We used the Transformer package~\cite{Wolf2019HuggingFacesTS} for training. The T5 model was trained for 12 epochs on a TPU using the default parameters from the T5 repository (learning rate 1e-3 with AdaFactor, batch size 8).\footnote{\url{https://github.com/google-research/text-to-text-transfer-transformer}} 
We truncate the summaries (and descriptions) to satisfy model-specific maximum input length.


\vspace{0.2cm}
\noindent\textbf{Results.}\space\space Table~\ref{tab:disc-res} shows the accuracies of different baselines. The highest accuracy is achieved by ALBERT-XXL ($83.33\%$) followed by T5-11B ($80.16\%$). Although both ALBERT and T5 were given partial character descriptions, 
their specific pre-training loss and larger number of parameters (for T5-11B) lead to superior performance over other baselines.
We observe that there is still a significant difference between the human performance ($91.80\%$) and the best model performance ($83.33\%$) on the character identification task, warranting future work on this direction.

\begin{table}[t]
\scriptsize
    \centering

\setlength\tabcolsep{2pt}
    \renewcommand{\arraystretch}{1.4}
    \begin{tabular}{lcccc p{0.001cm} c}
    \toprule
    \textbf{Model} & \textbf{BLEU} & \textbf{ROUGE-1} & \textbf{ROUGE-2} & \textbf{ROUGE-L} & & \textbf{BERT-F1} \\ 
    \midrule
    \rowcolor{blue!10} \multicolumn{7}{c}{\textbf{Length Truncated Input}} \\
    GPT2-L & 0.67 & 19.25 & 3.50 & 17.51 & & 77.71 \\ 
    BART-L & \textbf{1.38}  & \textbf{24.93} & \textbf{5.42} & \textbf{21.99} & & 84.54 \\
    Longformer & 1.05 & 21.47 & 4.66 & 19.37 & & 84.64 \\
    \midrule
    \rowcolor{blue!10} \multicolumn{7}{c}{\textbf{Coref Truncated Input}} \\
    GPT2-L & 0.58 & 18.69 & 3.15 & 16.91 & & 78.46 \\ 
    BART-L & 0.96 & 21.33 & 4.66 & 19.04 & & 84.26 \\
    Longformer & 0.98 & 21.18 & 4.40 & 19.13 & & 84.59 \\
    \midrule
    \rowcolor{blue!10} \multicolumn{7}{c}{\textbf{Full Length Input}} \\
    Longformer & 1.14 & 21.79 & 4.88 & 19.60 & & \textbf{84.72} \\
    \bottomrule
    \end{tabular}
\setlength\tabcolsep{6pt}
    \caption{Automatic evaluation results for \textit{Character Description Generation}. BART-L achieved the best BLEU and ROUGE scores while Longformer performed best on BERTScore. }
    \label{tab:gen-results}
\end{table}

\section{Character Description Generation}
\label{generative}



We present several strong baselines for generating character descriptions by fine-tuning pre-trained transformer-based language models (LM)~\cite{Vaswani:17}. We study two types of models: (1) a standard left-to-right LM, namely GPT2-L \cite{radford2019language} which is trained with LM objective to predict the next word; and (2) two encoder-decoder models, namely BART\footnote{We use the bart-large-xsum as initial weights as our task can benefit from the summarization capability.} \cite{lewis2019bart} and Longformer~\cite{Beltagy2020LongformerTL}\footnote{\url{https://github.com/allenai/longformer}. We initialize parameters of Longformer with the same pre-trained BART.} which initialize the state of the Transformer by reading the input, and learn to generate the output.  

One of the challenges of the proposed task is the length of the summaries, which might exceed the maximum allowable length for most existing pre-trained models. To overcome this, we 
either: (1) simply truncate the literature summary at the end, or (2) only keep sentences from the literature summary that have a mention of the character of interest. For the latter, we use a coreference resolution model, SpanBERT~\cite{joshi2019coref,spanbert}, 
to identify character mentions within a summary. This results in a modified dataset of character-specific literature summaries paired with character descriptions. In addition to these two approaches, we also fine-tune Longformer~\cite{Beltagy2020LongformerTL}
with original full-length literature summary. Longformer leverages an efficient encoding mechanism to avoid the quadratic memory growth and has been previously explored for NLU tasks (encoder-only). We integrate this approach into the pre-trained encoder-decoder BART model to encode inputs longer than its maximum token limit.
All the models take \texttt{[name] $c$ [sum] $S$ [desc]} as input and generate the character description $D^c$ as output.

\vspace{0.05cm}
\noindent\textbf{Experiment with Full Literary Pieces.} \space\space We also run an experiment on a subset of our data with accompanying full-text of the literary pieces.
Since it is infeasible to use the full texts as input given the memory constraints of current models, we coarsely select spans of the full-text 
beginning $50$ tokens before, and $50$ tokens after the occurrence of character's name. We use a Longformer model where the input is simply the concatenation of the 
selected spans.  Due to the small size of the this subset, we perform a $5$-fold cross validation starting from a pre-trained model fine-tuned on summary-description pairs.\footnote{Pre-training data do not contain instances of this subset.}



\vspace{0.05cm}

\noindent\textbf{Implementation Details.}\space\space We use the Transformer library~\cite{Wolf2019HuggingFacesTS}. Each baseline was trained for $5$ epochs with effective batch size of $8$, and initial learning rate of 5e-6. We use the maximum input length of $1024$ for GPT2, and $2048$ for BART\footnote{BART originally accepts inputs of maximum 1024 BPE-tokens. We extend this to 2048 by adjusting its positional embeddings.} and the variant of Longformer with truncated input. For experiment with original books, we use $16,384$ which is the maximum allowable input length for Longformer. During inference, we use beam search decoding with $5$ beams. 

\begin{figure}
    \centering
    \includegraphics[width=1.0\columnwidth]{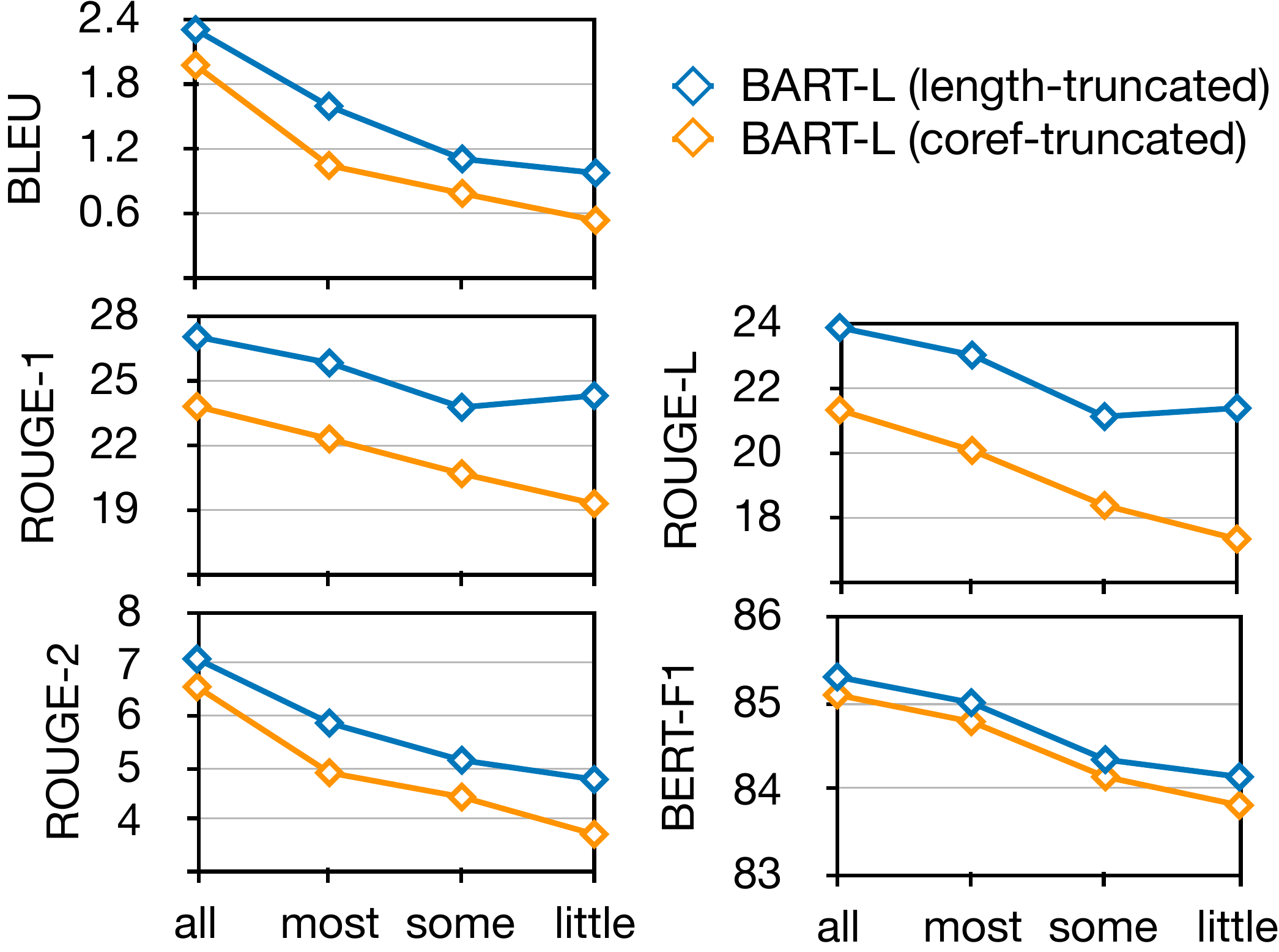}
    \caption{Breakdown results for BART-L on subsets with annotated fact coverage as all/most/some/little. Results for other baselines are provided in Appendix.}
    \label{tab:gen-results-fine}
\end{figure}

\subsection{Automatic Evaluation} 
Following previous works, we use several standard, widely used 
automatic evaluation metrics. We use \textbf{BLEU-4}~\cite{papineni2002bleu} that measures overlap of \textit{n}-gram up to $n=4$, \textbf{ROUGE-\textit{n}} ($n{=}1,2$), and \textbf{ROUGE-L} F-1 scores~\cite{lin2004rouge}\footnote{Note that we did not include perplexity score as it is not comparable across LM-based and encoder-decoder models.} 
. However, recent works~\cite{novikova2017we, wang-etal-2018-metrics} have raised concerns on the usage of these metrics as they fail to capture paraphrases and conceptual information. To overcome these issues, we additionally include a model-based metric, \textbf{BERTScore}~\cite{bert-score}, which measures the cosine similarity between contextualized embeddings of the gold and generated outputs.\footnote{We use the code at \url{https://github.com/Tiiiger/bert_score}}

The result of the automatic evaluation is presented in Table~\ref{tab:gen-results}. According to the table, BART-L consistently achieves the best performance across BLEU and ROUGE scores. However, Longformer achieves a slightly better BERTScore. 
Both BART and Longformer outperform GPT2 in general. This can be in part because BART and Longformer can handle longer context, and are initially pre-trained on a combination of books and Wikipedia data and further fine-tuned on summarization tasks, while GPT2 is pre-trained on WebText only.\footnote{While these models could have had access to the original book text, they do not have access to the character descriptions (our outputs) during pre-training. So, this information should not principally change any of our empirical conclusions.}  

Models perform relatively better in the length truncation setups than in the coreference truncation
. We posit that this is because a lot of the key points about major characters are likely to appear earlier in the book summary (favoring length truncation). Also, there might be errors introduced by the coreference resolution model itself.

\begin{table}[t]
\scriptsize
    \centering

\setlength\tabcolsep{2pt}
    \renewcommand{\arraystretch}{1.4}
    \begin{tabular}{lc p{0.0001cm} c p{0.0001cm} c}
    \toprule
    \textbf{Model} & \textbf{BLEU} & & \textbf{R-1}/ \textbf{R-2} /\textbf{R-L} & & \textbf{BERT-F1} \\ 
    \midrule
    Longformer $\small$ (w/ Books) & 0.73 & & 17.61 /3.60 / 16.15 & & 84.33 \\
    Longformer $\small$ (w/ Summaries) & \textbf{1.00} & & \textbf{19.46 / 4.33 / 17.74 } & & \textbf{84.77} \\
    \bottomrule
    \end{tabular}
\setlength\tabcolsep{6pt}
    \caption{Automatic evaluation results for models using full-text of books vs. literature summaries.}
    \label{tab:full-book}
\end{table}


In order 
to have a better insight into the models' performance with respect to varying level of task feasibility, in Fig.~\ref{tab:gen-results-fine}, we 
additionally report the breakdown of the results for BART-L on separate subsets 
with ``almost all'', ``most'', ``some'', ``little or none'' of the information about the character 
(refer to \textit{Fact Coverage} in \S\ref{data-validation}). As expected, we observe a consistent decline in the performance with lower amount of fact coverage. Results for other baselines are reported in Table~\ref{tab:gen-results-fine-rest} of the Appendix. 

In Table ~\ref{tab:full-book}, we compare the models when using selected spans from the original literary piece as the input vs. literature summaries as the input. 
We observe a decline in performance when we used the full text. 
This reveals that even though the literary pieces contain all the character information, this information is scattered which makes it harder 
for the model to identify important facts about the character.  Using full texts also requires encoders which are better at understanding dialog, first-person narratives and different writing styles of the authors.
We invite the community to consider this challenging but important problem.


\begin{figure}[t]
\centering
\includegraphics[width=0.90\columnwidth]{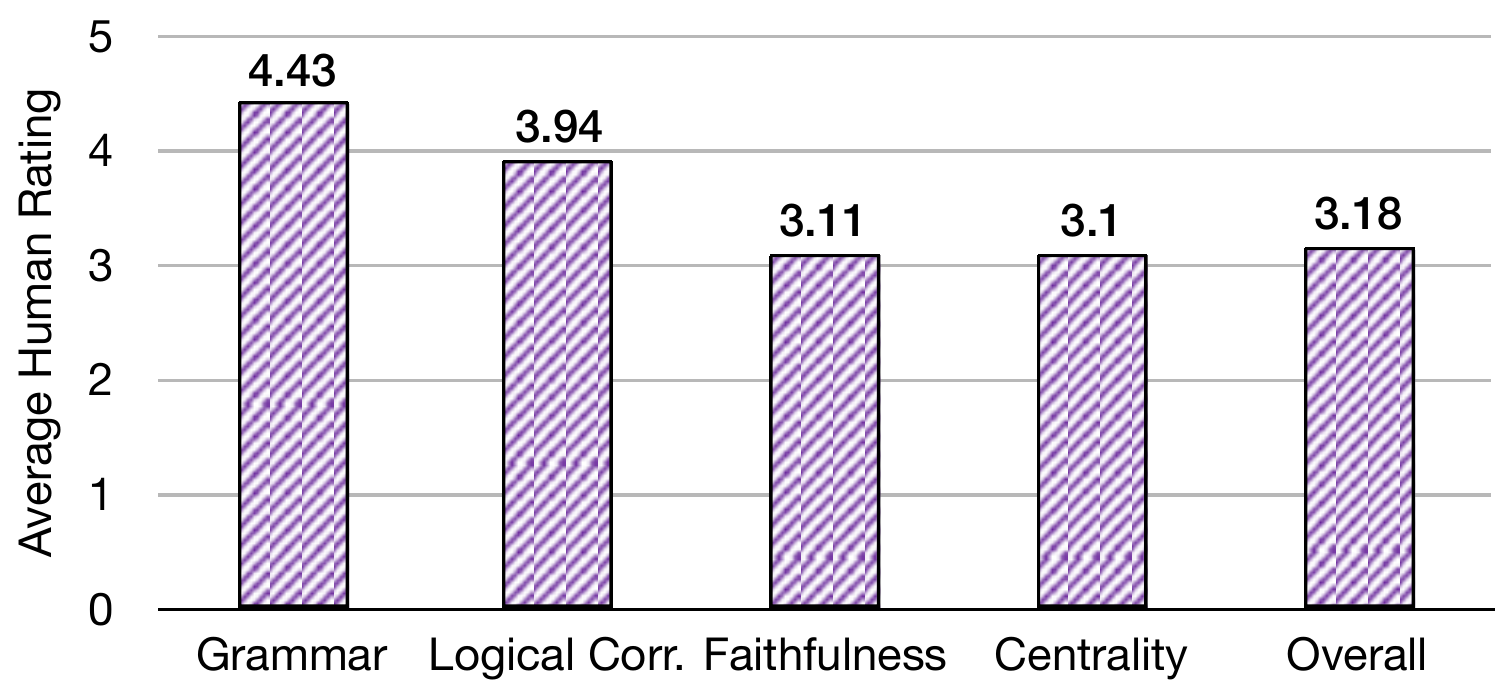} 
\caption{Human evaluation of generated character descriptions. While the descriptions are grammatically correct and logically coherent, they often misrepresent or miss important details about the character.}
\label{fig:human-desc-eval}
\end{figure}


\subsection{Human Evaluation} 
\label{sec:human_eval} 
To better evaluate the quality of the generated character descriptions, we conduct a human evaluation on $100$ test pairs of literature summaries and character descriptions generated by the BART-L model 
on Amazon Mechanical Turk.\footnote{Here we are evaluating $4$ character descriptions per summary, for the total of $25$ literature summaries.} Given a literature summary and multiple generated character descriptions (shown one by one), the workers were asked to rate each generated description on a Likert scale of $1-5$ ($1$ being the worst, and $5$ being the best) according to the following criteria: (1) \textbf{Grammatical correctness} to indicate if the generated description is grammatically correct, (2) \textbf{Logical correctness} to indicate whether the generated description is logically meaningful and coherent, (3) \textbf{Faithfulness} of the generated description with respect to the given summary (a faithful character description will not mention facts which are irrelevant to 
the character and/or not stated in the summary), (4) \textbf{Centrality} to evaluate whether the description captures important details and key facts about the character, and finally (5) the \textbf{Overall score} considering all the four criteria listed above. We provide a screenshot of the experiment in Fig.~\ref{fig:human-assess-2} of the Appendix. 

Fig.~\ref{fig:human-desc-eval} presents the results of this human evaluation. 
We observe that the generated descriptions show a reasonable level of grammatical ($4.43$) and logical correctness ($3.94$). However,  they lack behind when it comes to faithfulness ($3.11$) and centrality ($3.10$). 
We also report the distribution of ratings in Table~\ref{tab:human-eval-dist}. 
These results indicate that solving this task requires designing better models of character-centric analysis of narrative.

\begin{table}[t]
\footnotesize
    \centering
    \setlength{\tabcolsep}{4.7pt}
\renewcommand{\arraystretch}{1.2}
\begin{tabular}{lccccc}
\toprule
\textbf{Aspects} & \textbf{(1)} & \textbf{(2)} & \textbf{(3)} & \textbf{(4)} & \textbf{(5)}\\
\midrule


\textbf{Grammar} & \cellcolor{green!0.0} 0.00 & \cellcolor{green!5.5} 3.67 & \cellcolor{green!13.0} 8.67 & \cellcolor{green!43.0} 28.67 & \cellcolor{green!88.5} 59.00 \\
\textbf{Logical Corr.} & \cellcolor{green!2.5} 1.67 & \cellcolor{green!13.5} 9.00 & \cellcolor{green!28.5} 19.00 & \cellcolor{green!50.5} 33.67 & \cellcolor{green!55.0} 36.67 \\
\textbf{Faithfulness} & \cellcolor{green!19.0} 12.67 & \cellcolor{green!35.5} 23.67 & \cellcolor{green!31.5} 21.00 & \cellcolor{green!37.0} 24.67 & \cellcolor{green!27.0} 18.00 \\
\textbf{Centrality} & \cellcolor{green!23.0} 15.33 & \cellcolor{green!25.5} 17.00 & \cellcolor{green!40.5} 27.00 & \cellcolor{green!35.5} 23.67 & \cellcolor{green!25.5} 17.00 \\
\textbf{Overall} & \cellcolor{green!16.5} 11.00 & \cellcolor{green!29.0} 19.33 & \cellcolor{green!40.0} 26.67 & \cellcolor{green!40.0} 26.67 & \cellcolor{green!24.5} 16.33 \\

\bottomrule
\end{tabular}
    \caption{Percentage of different ratings from human evaluation of generated descriptions (1=worst, 5=best).}
    \label{tab:human-eval-dist}
\end{table}

\begin{table}[t]
\footnotesize
    \centering

\renewcommand{\arraystretch}{1.2}
\begin{tabular}{lc}
\toprule

\textbf{Error Type} & \textbf{Percentage} \\
\midrule
\textbf{Events} & \cellcolor{red!23.0} 46.00 \\
\textbf{Role} & \cellcolor{red!12.165} 24.33\\
\textbf{Relationships} & \cellcolor{red!12.5} 25.00 \\
\textbf{Personal characteristics} &  \cellcolor{red!6.165} 12.33 \\
\textbf{Behavioral characteristics} & \cellcolor{red!11.165} 22.33 \\
\textbf{No major error} & \cellcolor{red!13.835} 27.67\\
\bottomrule
\end{tabular}
    \caption{Error Analysis: proportion of generated descriptions with different error types.}
    \label{tab:analysis}
\end{table}

\subsection{Qualitative Analysis}
\label{sec:analysis}
\begin{table*}[t]
\footnotesize
    \centering


\definecolor{MengGreen}{rgb}{0.0, 0.5, 0.0}

\footnotesize
\setlength{\extrarowheight}{6pt}
\begin{tabular}{p{11cm}|p{3cm}}
\toprule
\textbf{Generated Character Descriptions} &  \textbf{Comments} \\ 
\toprule

\textbf{Book title: }The Three Sisters \\
\textbf{Character name:} Vershinin \\
\textbf{Generated Character Description:} Vershinin is one of the most interesting characters in Three Sisters. He's a \textcolor{MengGreen}{young officer in the Russian army}, and \textcolor{red}{he's got a lot of time on his hands.} But that doesn't stop him from having an affair \textcolor{red}{with Irina}, the youngest sister, who falls in love with him because she's \textcolor{MengGreen}{attracted to his philosophical ideas about the meaning of life.}  &  According to the summary ``Masha'', the middle sister, is the one who has an affair with the officer Vershinin, and not the youngest sister, Irina. \\ \midrule

\textbf{Book title:} The Sun Also Rises \\
\textbf{Character name:} Pedro Romero\\
\textbf{Generated Character Description:} Pedro Romero is \textcolor{MengGreen}{a young Spanish bull-fighter who is in love with Lady Brett Ashley, a beautiful British divorcee.} He's a talented and \textcolor{MengGreen}{talented young man}, and, despite the fact that he is only \textcolor{red}{twenty-five years old}, he has \textcolor{red}{the potential to be one of the best bullfighters in Pamplona}. \textcolor{MengGreen}{Jake feels terrible for introducing him to Brett, fearing that it has corrupted him.}  &  The description captures most of the important details about the character. \\ \midrule

\end{tabular}

\vspace{-1em}

    \caption{Examples of generated descriptions. Words in \textcolor{red}{red} correspond to hallucinated or missing content, and words in \textcolor{MengGreen}{green} correspond to faithful information. The input literature summaries are provided in the Appendix. }
    \label{tab:qual-example}
\end{table*}

Here, we do a qualitative analysis for the \textit{Character Description Generation} task. In our human evaluation of the generated character descriptions (\S\ref{sec:human_eval}), we additionally provided a questionnaire to collect in-depth feedback from crowd-workers on the type of errors the BART-L model made when generating character descriptions from the given literature summaries. The questionnaire asked ``What details about the character does the given character description miss or describe inaccurately. Note that the description is supposed to describe only the important details and not necessarily all of them.'' The workers were asked to select all the applicable choices among the following error types: 

\begin{enumerate}[noitemsep] 
\item \textbf{Events: }The character description misses or misrepresents some main event(s) that the character is involved in.
\item \textbf{Role: }The character's role in the narrative (e.g., protagonist, antagonist, etc.) is important but is not included or misrepresented in the character description. 
\item \textbf{Relationships: }The character's relationship with other characters is important (e.g., the protagonist's wife) but is not included or misrepresented in the character description. 
\item \textbf{Personal characteristics: }The character's personal characteristics (e.g., age, ethnicity, personality, etc.) are important for the narrative but are not included or misrepresented in the character description. 
\item \textbf{Behavioral characteristics: }The character's motivation, desires, 
and behavior are important but are not included or misrepresented in the character description. 
\item \textbf{No major error: }None of the above. The character description captures most of the important details about the character. 
\end{enumerate} 
We also provided an optional text box for them to type in other details that are missing or misrepresented but 
not listed above. 

The result of this analysis is shown in Table~\ref{tab:analysis}. 
We can see that the generated descriptions make fewer mistakes in capturing personality-related attributes ($12.33\%)$ and more mistakes in representing important events involving the characters ($46\%$). They also sometimes omit or misrepresent  roles ($24\%$),  relationships ($25\%$), and behavioral characteristics ($22\%$) of the characters. This indicates  factors that future systems should consider improving upon when addressing this task. 


We provide qualitative examples of the generated character descriptions along with the errors they made (as pointed out by the turkers) in Table~\ref{tab:qual-example}.
More examples with input literature summaries are provided in Tables~\ref{tab:errore_examples1} to \ref{tab:errore_examples4} of the Appendix.



\section{Conclusion}
\label{sec:conclusion}

Understanding and critically analyzing fictional characters is an important element of understanding a literary piece.
Human readers build a mental model of characters, understand what they look like, their role in the literary piece, and assess their psychology, motivations, and consequences of their behavior. However, building such a deep understanding of fictional characters in narratives is hard for machine reading systems.
To encourage progress in character-centric understanding of narratives, we present \fiscu, a dataset of literature summaries paired with descriptions of characters that appear in them. We use \fiscu to propose two tasks that 
explore the ability of the modern neural models to understand the narrative from the perspective of characters.
Performing human assessments on the model outputs show that there is still a lot of room for improvement on these tasks.

\section*{Acknowledgments}
This work was supported in part by ETH Grant (ETH-19 21-1) and NSF grant IIS2047232. We would also like to thank Jena D. Hwang for helping with designing the AMT task.

\section*{Broader Impacts and Ethics Statement}
\label{sec:bias}

\noindent{\bf Bias in Narrative Texts:}
\fiscu is based on novels which often reflect societal norms and biases of their times. Such a dataset can be used to understand societal bias as well as design Natural Language Understanding models that can be more aware of and possibly even avoid such biases. With this motivation, we analyzed the issue of gender bias in \fiscu.

First, we inferred the gender of the characters in our dataset using the pronouns used to refer to them. We could not infer the gender of some of the characters because of errors in the coreference system or lack of enough mentions, and we filtered them out for this analysis. We found that there are significantly more male characters than female characters in our dataset. Specifically, $66\%$ of the characters are male. This suggests that systems that do not account for this bias might end up having more training data (and hence yield better performance) on descriptions of male characters than of female characters. 

Second, we also investigated the scope of gender bias in the summaries. We computed the average number of mentions of male and female characters (in the summaries). We found that on average male and female characters are mentioned $32.1$ and $31.7$ times, respectively. This indicates that even though there are fewer female characters in the literary pieces of our dataset, the ones that are present play a significant role in the development of the narrative. Possibly because of their importance in the narrative, they are mentioned as many times as male characters in the summary (which describes the main developments and not all details from the literary piece). 

Third, we investigated if the literary experts who composed the descriptions were biased in their analysis. For this, we compute the length of character descriptions of various characters. We found that there is no significant difference between male and female characters in this aspect. Specifically, the average number of tokens in the description of a male character was $203$, and that of a female character was $200$. Also, the average number of sentences in the description of a male character was $9.4$ and that of a female character was $9.3$. This also aligns with our observation in the previous experiment where we found that female characters, though fewer, play important roles in the narrative, and so their descriptions are not any shorter than descriptions of male characters. Overall, this analysis suggests that descriptions are not biased in their treatment of male and female characters.    

In any language generation setting, such as ours, there is the possibility of (potentially harmful) social biases that can be introduced in the training data. As we did not specifically control or regularize our model to remove the possibility of such biases, we would urge downstream users to undertake the necessary quality-assurance testing to evaluate the extent to which such biases might be present and impacting their trained system and to make modifications to their model and procedures accordingly. 

\noindent{\bf Human participation in our study :} We conducted 2 human evaluations on Amazon Mechanical Turk. To ensure the annotators were fairly compensated, we did several rounds of test runs and estimated the average time to finish one HIT. Workers were paid \$12/hr based on the HIT timings.
We did not ask any personal, sensitive or identifying information from the annotators.




\bibliography{naaclhlt2019}
\bibliographystyle{acl_natbib}

\appendix

\section{Collecting Annotations from Crowd Workers}
\label{appendix-annotations}
To alleviate the limitations of crowd-sourcing and ensure high quality of annotations, we took several steps. First, we conducted a pilot annotation exercise  where we (authors) assessed the feasibility of the proposed task on a subset (250 instances) of the data. This pilot annotation helped us set up the task on AMT in a way that would make the task feasible for turkers (e.g. by asking clear concise questions).  Second, we designed our setup to avoid annotator fatigue by asking them to read the summary context once and answer questions about all characters in that summary. Third, we ran a few experiments on AMT (before annotating the entire set) where we also included a `comment` section for turkers to allow them to bring up issues or ambiguities in our setup. We then manually analyzed the results and modified the tasks based on the comments. Finally, after annotating the entire set, we computed inter-annotator agreement as a way to ensure trust in the annotation quality. We found reasonable agreements between annotators as reported in Footnote 10 of the paper. We would also like to mention that we received several comments from the annotators that they found the task very interesting and enjoyable.

\begin{table*}[t]
\scriptsize
    \centering
\setlength\tabcolsep{3.3pt}
    \renewcommand{\arraystretch}{1.4}
    \begin{tabular}{lc p{0.001cm} ccc p{0.001cm} c}
    \toprule
    \textbf{Model} & \textbf{BLEU} && \textbf{ROUGE-1} & \textbf{ROUGE-2} & \textbf{ROUGE-L} & & \textbf{BERT-F1} \\ 
    \midrule
    \rowcolor{blue!10} \multicolumn{8}{c}{\textbf{Length Truncated Input}} \\
    \textbf{GPT2-L} & 0.90/0.71/0.60/0.59 & & 20.33/19.94/18.81/19.39 & 4.34/3.60/3.36/3.32 & 18.50/17.91/17.14/17.84 & & 76.23/80.11/76.53/75.81 \\ 
    \textbf{Longformer} & 1.86/1.06/0.92/0.70 & & 24.16/21.78/20.55/20.20 & 6.80/4.62/4.33/3.98 & 22.05/19.65/18.71/18.26 & & 85.60/84.92/84.46/84.24 \\
    \midrule
    \rowcolor{blue!10} \multicolumn{8}{c}{\textbf{Coreference Truncated Input}} \\
    \textbf{GPT2-L} & 0.82/0.63/0.53/0.58 & & 19.80/19.24/17.96/18.49 & 3.86/3.24/3.06/3.04 & 17.79/17.39/16.46/16.62 & & 76.16/79.52/78.33/80.23 \\ 
    \textbf{Longformer} & 1.78/1.09/0.77/0.65 & & 23.32/22.23/20.23/19.90 & 6.04/4.80/3.96/3.57 & 21.41/20.22/18.16/17.70 & & 85.43/85.07/84.47/84.12\\
    \midrule
    \rowcolor{blue!10} \multicolumn{8}{c}{\textbf{Full Length Input}} \\
    \textbf{Longformer} & 2.15/1.31/1.04/0.65 & & 24.47/22.56/20.90/20.63 & 6.98/5.37/4.63/3.84 & 21.91/20.40/18.85/18.48 & & 85.66/85.11/84.57/84.35 \\
    \bottomrule
    \end{tabular}
\setlength\tabcolsep{6pt}
    \caption{Breakdown results on subsets of test set with annotated fact coverage as all/most/some/little.}
    \label{tab:gen-results-fine-rest}
\end{table*}


\begin{table*}[t]
    \renewcommand{\arraystretch}{1.2}
    \scriptsize
    \centering
    
\footnotesize
\begin{tabular}{p{12cm}|p{3cm}}
\toprule
\textbf{Generated Description} &  \textbf{Comments} \\ 
\toprule
\textbf{Book title: }The Three Sisters\\
\textbf{Character name:} Vershinin\\
\textbf{Summary:} Three Sisters mainly follows the story of--wait for it--three sisters: Olga, Masha, and Irina Prozorov. They live with their brother, Andrey, in a big house on the edge of a small Russian town. The townspeople are kinda backward and boring compared to their educated and culture-lovin' family, so this set of sibs is not too fond of the town to begin with. Believe it or not, the only halfway interesting people around are the guys in the military. Basically, the Prozorov kids are worldly, well-educated army brats. And being in the army in Tsarist Russia pretty much meant you were in with the aristocracy and, once you got through the fighting stuff, probably developed a taste for the finer things in life. So ever since the family moved from Moscow eleven years prior (with their father, now dead), the sisters have obsessed over the dream of moving back to the big city. But guess what? It's not happening. Olga, the eldest, is a spinster schoolteacher and eventually becomes a headmistress living with her elderly maid. Masha, the middle sister, is married to another schoolteacher, Kulygin, whom she despises for his small-mindedness. She has an affair with the officer Vershinin because he's given to just the sort of philosophizing that really starts her engines. And guess what happens: the affair ends in heartbreak. Poor Irina, the youngest, has fanciful ideas about the value of work, but soon realizes that, in reality, work sucks the life out of her. She's also in love with the idea of love, but doesn't get to experience it. Finally she comes around to saying "yes" to Baron Tuzenbach, a friendly but ugly man who's been after her for years. On the day they're leaving to get married, he gets shot in a duel. Bummer. Andrey, the brother, gives up his intellectual dreams to pursue a town girl, Natasha. They marry, have kids, and little by little she takes over the estate. At the end of the play, the upper-class Prozorovs are pretty much evicted from their own house, while Natasha, a symbol of the working class, is on the rise. Allegory, much? For anyone not already reaching for the history books, this was just a few years before the Bolshevik Revolution of 1917--and you better believe smart Ruskies like Chekhov could already sense some storms on the horizon. &  According to the summary ``Masha'', the middle sister, is the one who has an affair with the officer Vershinin, and not the youngest sister, Irina.  \\
\textbf{Generated Character Description:} Vershinin is one of the most interesting characters in Three Sisters. He's a \textcolor{MengGreen}{young officer in the Russian army}, and \textcolor{red}{he's got a lot of time on his hands.} But that doesn't stop him from having an affair \textcolor{red}{with Irina}, the youngest sister, who falls in love with him because she's \textcolor{MengGreen}{attracted to his philosophical ideas about the meaning of life.}  \\ \midrule

\end{tabular}

    \caption{Qualitative example 1 for the generated descriptions. Words in \textcolor{red}{red} correspond to hallucinated or missing content, words in \textcolor{MengGreen}{green} correspond to faithful information. }
    \label{tab:errore_examples1}
\end{table*}

\begin{table*}[t]
    \renewcommand{\arraystretch}{1.2}
    \scriptsize
    \centering
    
\footnotesize
\begin{tabular}{p{12cm}|p{3cm}}
\toprule
\textbf{Generated Description} &  \textbf{Comments} \\ 
\toprule
\textbf{Book title:} The Sun Also Rises \\
\textbf{Character name:} Pedro Romero\\
\textbf{Summary}: Jake Barnes and his expatriate friends live in the topsy-turvy, hedonistic (sensual and self-indulgent) world of post-World War I Paris. There, they occasionally work, but spend most of their time partying, drinking, and arguing. From Jake's perspective, we meet the cast of characters that populates his story: the most important among them are Robert Cohn, a weak-willed, down-on-his-luck Princeton grad and unsuccessful writer, and Lady Brett Ashley, an exciting, beautiful, and unpredictable British divorcee. Although Jake and Brett are actually in love, they aren't together, presumably because a mysterious war wound has rendered Jake impotent. Cohn falls in love with Brett (as everyone does) and, despite the fact that she's not terribly impressed with him, she secretly goes on a trip with him to the Spanish resort town of San Sebastian. Cohn is infatuated with Brett--he's completely smitten. We're talking truly, madly, deeply in smit. Unfortunately for Cohn (and for everyone, for that matter), Brett is engaged to a wealthy, charming, and utterly inept drunkard named Mike. Jake's whimsical friend Bill returns to Paris from a trip and a plan is born: everyone agrees to decamp to Spain for some fishing and the running of the bulls in Pamplona. On their brief fishing trip, Bill and Jake have a splendid time communing with nature and with each other, but the relaxation quickly comes to an end. They return to civilization and meet up with Brett, Mike, and Cohn in Pamplona for a weeklong orgy of bullfights, alcohol, and high drama. Jake has a true passion (aficion) for bullfighting, but everyone else is simply there to have a good time. Brett begins a scandalous affair with a passionate and talented young bull-fighter, Pedro Romero. Jake feels terrible for many reasons--among them is the fear that he has corrupted Romero in some way by introducing him to Brett. Cohn's thwarted infatuation with Brett leads to arguments with everyone and, finally, he beats the unfortunate Romero to a bloody pulp. As the fiesta winds down, everyone leaves Pamplona in various states of anxiety, depression and frustration.Jake heads to San Sebastian, where he intends to decompress alone for a while. Unfortunately, desperate telegrams from Brett arrive immediately. He goes to her in Madrid, where she is alone, having sent Romero away. For the first time, we see Brett truly vulnerable, afraid, and guilty. The future looks just as bleak--Jake and Brett agree again that, even though they love each other, they can't be together.  &  The description captures most of the important details about the character.  \\
\textbf{Generated Character Description:} Pedro Romero is \textcolor{MengGreen}{a young Spanish bull-fighter who is in love with Lady Brett Ashley, a beautiful British divorcee.} He's a talented and \textcolor{MengGreen}{talented young man}, and, despite the fact that he is only \textcolor{red}{twenty-five years old}, he has \textcolor{red}{the potential to be one of the best bullfighters in Pamplona}. \textcolor{MengGreen}{Jake feels terrible for introducing him to Brett, fearing that it has corrupted him.}   \\ \midrule

\end{tabular}
    \caption{Qualitative example 2 for the generated descriptions. Words in \textcolor{red}{red} correspond to hallucinated or missing content, and words in \textcolor{MengGreen}{green} correspond to faithful information. }
    \label{tab:errore_examples2}
\end{table*}

\begin{table*}[t]
\renewcommand{\arraystretch}{1.2}
    \scriptsize
    \centering
    \footnotesize
\begin{tabular}{p{12cm}|p{3cm}}
\toprule
\textbf{Generated Description} &  \textbf{Comments} \\ 
\toprule
\textbf{Book title: }The Waves\\
\textbf{Character name:} Neville\\
\textbf{Summary:} The story begins by introducing us to the novel's six (yup, you read that right) narrators, Bernard, Neville, Louis, Jinny, Susan, and Rhoda, who meet as children in a nursery. During this phase of the novel, we learn a lot about the characters' personalities and their relationships to each other. After looking on as our new friends get embroiled in some kid-level dramas (e.g., trouble in math class and unrequited crushes), the six children head off to their respective boarding schools. At that time, the boys meet Percival, whom everyone seems to revere (and Neville falls in love with). The protagonists then all graduate and proceed into their adult careers (with a stop at university along the way, for some). At some point in there, Percival becomes friends with the girls as well, though we're not sure when that actually occurs. The narrators' paths diverge quite a bit as the novel progresses. After enduring a stint in a Swiss school, Susan returns to her beloved hometown, gets married, and starts having babies. Meanwhile, Bernard apparently moves to Waterloo (that's not entirely clear, but Woolf drops some clues to that effect), and we're not entirely sure what he does there, other than shave and make up stories about pedestrians. Meanwhile, Rhoda, Louis, Neville, and Jinny go to live in London, and their life paths are all over the map: Louis works for a shipping company, Jinny is a socialite, and Neville is a classics professor (we don't learn Rhoda's profession). Midway through the book, the friends meet up for dinner in London to see Percival off before he leaves to work in India, at which point Bernard announces that he is getting married. An unspecified amount of time later, Percival falls from his horse in India and dies, and our narrators are devastated. The death occurs just as Bernard's son is born, creating some serious cognitive dissonance for poor Bernard. The characters then struggle with aging and reflect upon the progress of their lives and adulthood. Toward the end, they all meet up at Hampton Court and experience the aches and pains that come when old friends reconnect after a long time has passed. The novel ends with Bernard talking to an apparent stranger, attempting to sum up the lives of the six narrators (i.e., the events of the novel) and work out his philosophies of language and lifeand that's why the last chapter is fifty pages long. He reflects upon his lifelong struggle to turn his phrasemaking into something meaningful and, despite the major ups and downs he's experienced along the way, he resolves to keep on trying. &  The model hallucinated some details such as being a professor at Oxford University while the summary only mentions: He lives in London and is a  classics professor.  Also, phrases such as ``We don't learn a whole lot about''  appears quite frequently in the generated descriptions.   \\ \par 
\textbf{Generated Character Description:} Neville is one of the six narrators. \textcolor{MengGreen}{He's a classics professor} at \textcolor{red}{Oxford University} and, \textcolor{red}{like the other five}, he's also a classicist. \underline{We don't learn a whole lot about} Neville, other than the fact that \textcolor{MengGreen}{he falls in love with Percival}, which makes him a bit of an oddball. In fact, we're not even sure what he does in his spare time away from Oxford.  \\
\midrule
\end{tabular}
    \caption{Qualitative example 3 for the generated descriptions. Words in \textcolor{red}{red} correspond to hallucinated or missing content, words in \textcolor{MengGreen}{green} correspond to faithful information, and \underline{underline} corresponds to generic repetitive content. }
    \label{tab:errore_examples3}
\end{table*}

\newpage

\begin{table*}[t]
\renewcommand{\arraystretch}{1.2}
    \scriptsize
    \centering


\definecolor{MengGreen}{rgb}{0.0, 0.5, 0.0}

\footnotesize
\begin{tabular}{p{12cm}|p{3cm}}
\toprule
\textbf{Generated Description} &  \textbf{Comments} \\ 
\toprule
\textbf{Book title: }Travels with Charley \\
\textbf{Character name:} Charley the Dog \\ \par
\textbf{Summary:} Because he's feeling pretty out of touch with his own country--and he's considered a great American author and all that--John Steinbeck decides to take a road trip around the U.S. to check it out and get a sense of where Americans and their hometowns are at in 1960. To get all prepped, he commissions a souped-up truck with a little house on the back that he can live in when he isn't crashing at hotels. He calls the truck "Rocinante" after Don Quixote's horse--clever, huh? When he's all set (and after a small run-in with a hurricane just before he was supposed to leave), he and Charley (his French poodle) hit the road. He starts out by driving over into Connecticut from his home in Long Island (with some assists from ferries, natch) and then heads north into New England. Along the way, he meets a pretty colorful group of characters and learns about their ways of life and their perspectives on the country and its politics. Also, he kind of takes the temperature of regional "temperaments" along the way. Then he comes back down out of New England and heads west, crossing through New York. He tries to cut through Canada, but he gets into a kerfuffle at the border because Charley doesn't have his proof of rabies vaccination, so he has to turn around. Steinbeck then passes through the Midwest, continuing to offer his reflections and thoughts about the people and places he encounters along the way. When he gets to Chicago, he puts Charley in a kennel and enjoys a couple of days with his wife, who flew out to meet him. He doesn't give us details of their time together, though. After that brief interlude, he heads further west into Minnesota and Wisconsin. He hits bad traffic and gets lost around the Twin Cities, and he's charmed by Wisconsin and its dells. He also visits Sauk Centre, the birthplace of author Sinclair Lewis. Then he heads toward Fargo, North Dakota, which apparently had been the subject of his boyhood fantasies. We picture Hawaii when we're fantasizing about faraway places, but okay... He heads through North Dakota and the Bad Lands, warming up to that area quite a bit when he sees it at sunset (otherwise, it's pretty scary, apparently). When he goes through Montana, he loves it. However, he soon has to head back south so he can see Yellowstone... though he doesn't stay long, since Charley freaks out about the bears. Then, he heads to Washington. Along the way, Charley ends up getting sick with a bladder problem, and Steinbeck tries to get him some help from a vet in Spokane. The doc isn't actually super-helpful, though. Still, they get the problem resolved enough that they can continue driving. Heading to Seattle, Steinbeck finds the place changed a lot from the last time he was there (and not for the better, sadly). Then he heads into Oregon, where his tire blows out. Through the kindness of a dude at the one service station open in that area on a Sunday, though, he gets some new heavy-duty tires and is on his way pretty quickly. Then he heads south through California, visiting Salinas (his hometown) on the way. While there, he argues with his sisters about politics and reconnects with old friends, but apparently it's kind of awkward, and he's reminded of that old adage that you can't go home again. After that, he starts circling back toward home, driving through the Mojave Desert and through the Southwest and Texas. While he's in Texas, he meets up with his wife once again and stays with some friends (oh, and actually takes Charley to a competent vet in Amarillo). We get a lot of detail about Steinbeck's impressions of Texas. After that little break, Steinbeck goes to Louisiana so he can see for himself what's going on there in terms of racial tensions (which appear to be pretty high). There was a well-known protest going on: two African American children had been admitted to a New Orleans school, and lots of people had a huge problem with that fact. So, some women known as the Cheerleaders basically stood outside every morning to scream at these poor children as they headed into school. Steinbeck went to get a firsthand look at this spectacle and was, as you might imagine, disgusted. He then headed back home, finally deciding he was "done" with the journey somewhere in Virginia.  & The description never mention anything about the \textcolor{red}{Charley's health struggles} as appeared in the summary we have: ``Along the way, Charley ends up getting sick with a bladder problem, and Steinbeck tries to get him some help from a vet in Spokane. The doc isn't actually super-helpful, though.'' \\ \par
  \textbf{Generated Character Description:} When we first meet Charley, \textcolor{MengGreen}{he's a French poodle. He's Steinbeck's pet and his companion for most of the road trip}, and we get to see him grow up a bit. In fact, \underline{we see a lot of him in the book}, from the time he first meets him until he gets to Chicago. We also get a glimpse of his personality as a character in his own right, when he tells us, ``He was a good-natured, loving dog'' (1.2.1).  \\ \midrule
\end{tabular}

    \caption{Qualitative example 4 for the generated descriptions. words in \textcolor{MengGreen}{green} correspond to faithful information, and \underline{underline} corresponds to generic repetitive content. }
    \label{tab:errore_examples4}
\end{table*}


\begin{figure*}[t]
\centering
\includegraphics[width=0.9\linewidth]{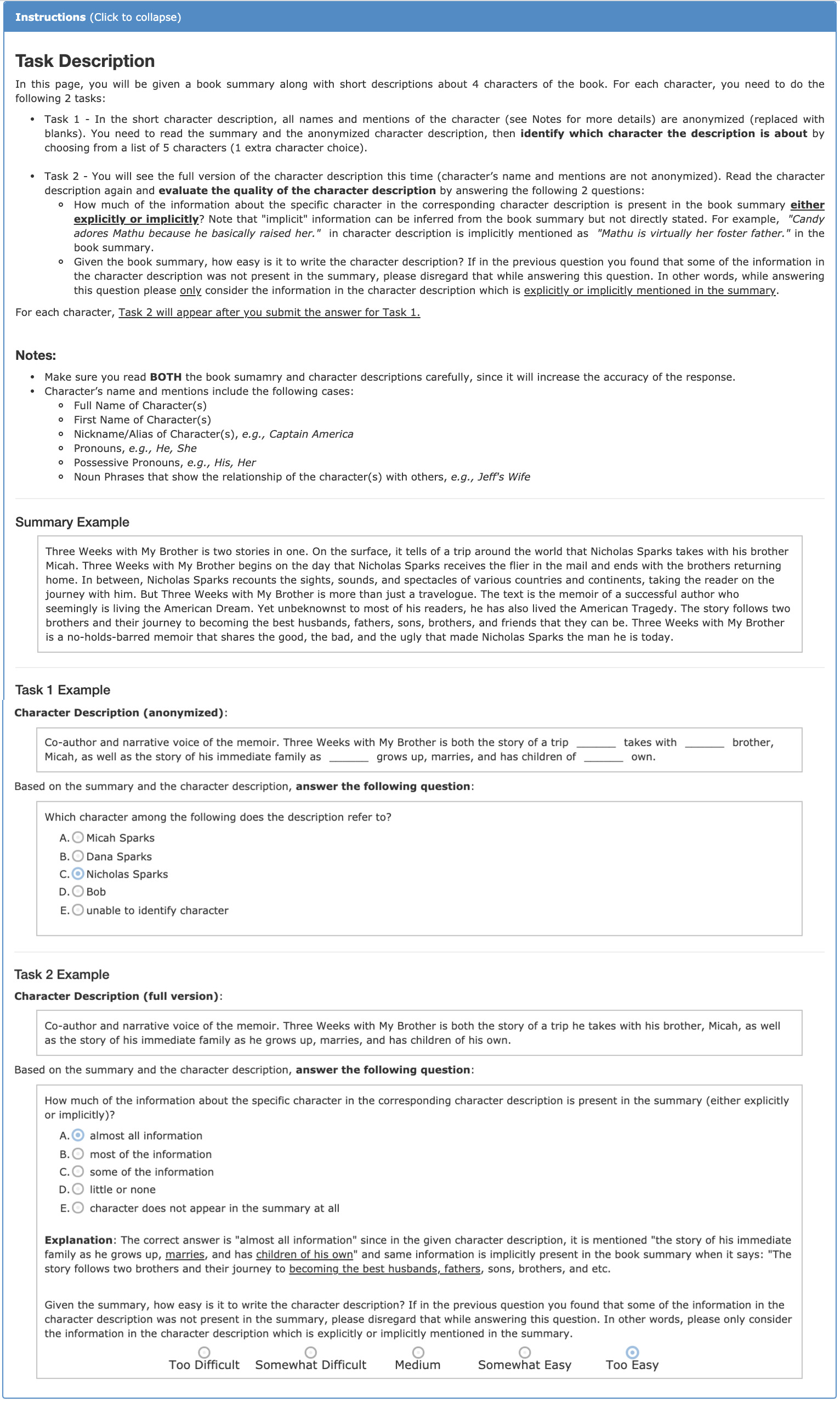}
\caption{An illustration of human assessment on AMT.}
\label{fig:human-assess-1}
\end{figure*}

\begin{figure*}[t]
\centering
\includegraphics[width=0.8\linewidth]{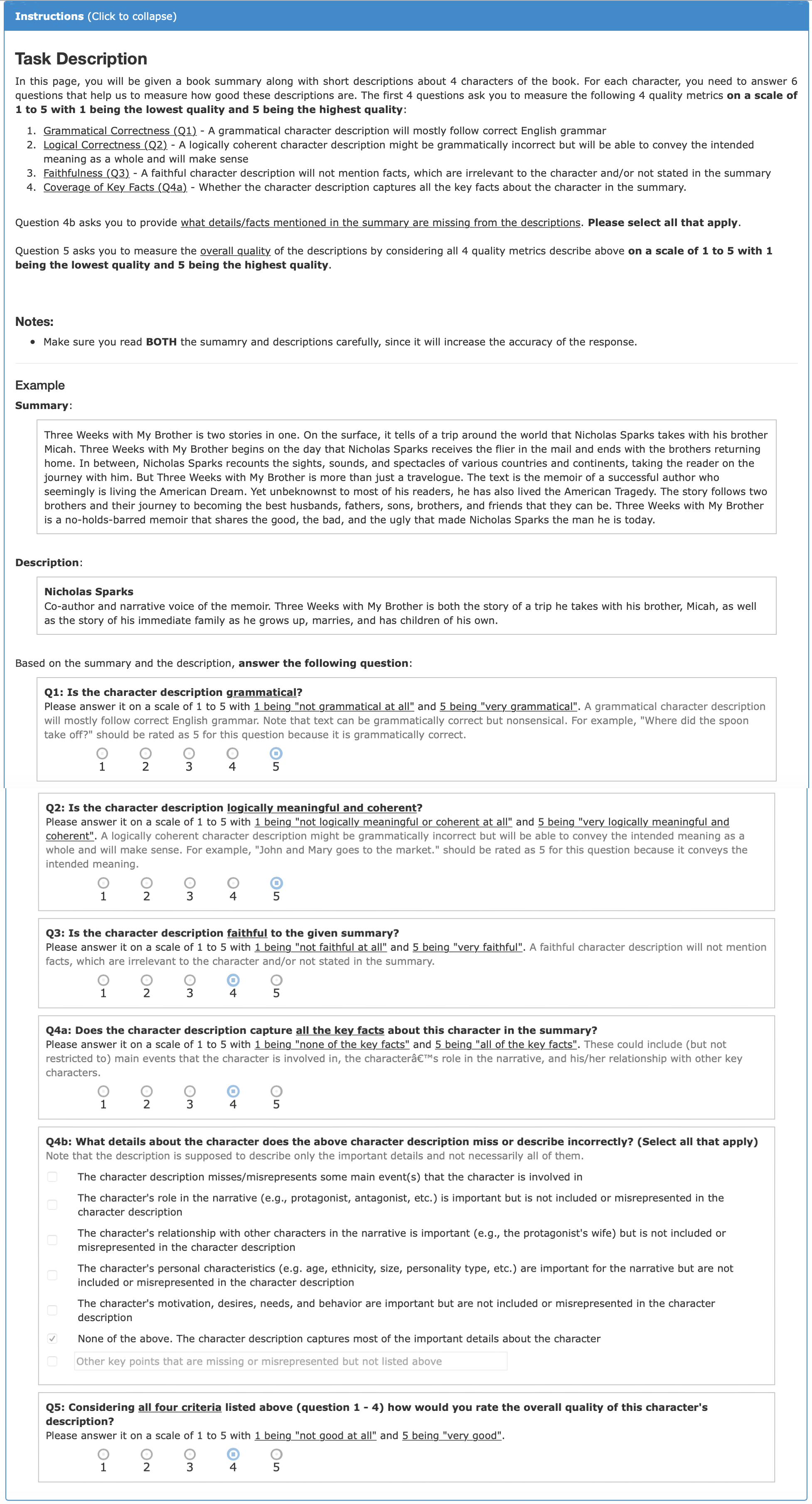}
\caption{An illustration of human evaluation for generated character description.}
\label{fig:human-assess-2}
\end{figure*}


\end{document}